\documentclass[letterpaper, 10 pt, journal, twoside]{ieeetran}
\usepackage{times}
\usepackage{graphics} %
\usepackage{graphicx}
\usepackage{subfigure}
\usepackage{amsmath,amssymb,amsopn,amstext,amsfonts}
\usepackage{cancel}
\usepackage[space]{cite}
\usepackage{pdfsync}
\usepackage{balance}
\usepackage{color}
\usepackage{mathtools}
\usepackage{bm}

\usepackage{diagbox}
\usepackage{float}
\usepackage{epstopdf}
\usepackage{pifont}
\usepackage{fixltx2e}
\usepackage{amsmath}
\usepackage{multirow}
\usepackage{url}
\usepackage{verbatim}
\usepackage{caption}
\usepackage{adjustbox}
\usepackage{booktabs}
\usepackage{threeparttable}
\usepackage{makecell}
\usepackage{upgreek}
\usepackage{algorithm, algpseudocode}

\newcommand{\titlefull}[0]{NeuroVE: Brain-inspired Linear-Angular Velocity Estimation with Spiking Neural Networks}

\makeatletter
\let\NAT@parse\undefined
\makeatother
\usepackage[linkcolor=red,citecolor=green,urlcolor=red,colorlinks=true]{hyperref}

\begin{document}
\title{\titlefull}

\author{Xiao Li$^{1,2}$, Xieyuanli Chen$^{1}$, Ruibin Guo$^{1}$, Yujie Wu$^{3}$, Zongtan Zhou$^{1}$, Fangwen Yu$^{2}$, Huimin Lu$^{1}$%
\thanks{${1}$ College of Intelligence Science and Technology, National University of Defense Technology. {\tt\small lhmnew@nudt.edu.cn} }
\thanks{${2}$ Center for Brain-Inspired Computing Research, and Department of Precision Instrument, Tsinghua University, Beijing 100084, China. {\tt\small yufangwen@tsinghua.edu.cn}}
\thanks{${3}$ Department of Computing, the Hong Kong Polytechnic University, Hong Kong SAR.}
\thanks{ 
Corresponding author: Huimin Lu, Fangwen Yu.}
}

\maketitle
{X. Li, \MakeLowercase{\textit{et al.}}: \titlefull}

\begin{abstract}

Vision-based ego-velocity estimation is a fundamental problem in robot state estimation.
However, the constraints of frame-based cameras, including motion blur and insufficient frame rates in dynamic settings, readily lead to the failure of conventional velocity estimation techniques.
Mammals exhibit a remarkable ability to accurately estimate their ego-velocity during aggressive movement. 
Hence, integrating this capability into robots shows great promise for addressing these challenges.
In this paper, we propose a brain-inspired framework for linear-angular velocity estimation, dubbed NeuroVE.
The NeuroVE framework employs an event camera to capture the motion information and implements spiking neural networks (SNNs) to simulate the brain's spatial cells' function for velocity estimation.
We formulate the velocity estimation as a time-series forecasting problem.
To this end, we design an Astrocyte Leaky Integrate-and-Fire (ALIF) neuron model to encode continuous values. 
Additionally, we have developed an Astrocyte Spiking Long Short-term Memory (ASLSTM) structure, which significantly improves the time-series forecasting capabilities, enabling an accurate estimate of ego-velocity.
Results from both simulation and real-world experiments indicate that NeuroVE has achieved an approximate 60\% increase in accuracy compared to other SNN-based approaches.

\end{abstract}
\vspace{0.2cm}


\section{Introduction}
\label{sec: introduction}

\IEEEPARstart{V}{ision-based} ego-velocity estimation is essential for robot state estimation, especially in dynamic environments. 
In intricate, highly dynamic scenes, conventional frame-based cameras encounter limitations such as motion blur and varying illumination.
These limitations often hinder the detection of visual features by traditional vision methods, potentially resulting in failures.

\begin{figure}[t]
    \centering
    \includegraphics[width=1\linewidth]{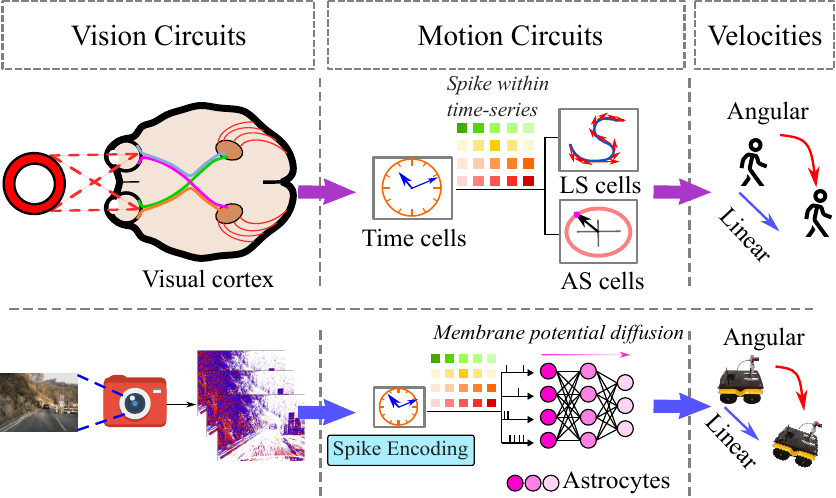}
    \caption{
    The linear and angular speed estimation pipeline mirrors the processes of the human brain and robotic systems. 
    In the vision circuits, the sensor transforms external light into spike signals. 
    In the motion circuits, these signals are first encoded with temporal information. 
    Subsequently, linear speed (LS) and angular speed (AS) cells translate these spikes into velocity signals.
    }
    \vspace{-0.5cm}
    \label{fig: pipeline}
\end{figure}

Humans' visual perception and motion estimation demonstrate impressive abilities, particularly in accurately estimating speed and responding swiftly.
In the human brain, there exists a special class of cells known as spatial cells, which include linear speed (LS) cells, angular speed (AS) cells~\cite{Spalla2022NC}, and time cells~\cite{Eichenbaum2014nrn}. %
As shown in Fig~\ref{fig: pipeline}, the vision motion circuits receive external light signals, which are converted into spikes and then conveyed to the cerebral cortex, where they activate spatial cells.
These essential discoveries have inspired us to solve the problem of velocity estimation by mimicking the vision motion circuits in the brain.

In terms of vision circuit modeling, the neuromorphic sensor~\cite{Yang2024nature}, e.g. event cameras~\cite{Gallego20pami}, has been developed by leveraging the principles of brain-based vision. 
It exhibits exceptional temporal resolution and dynamic range, enabling excellent performance in dynamic scenarios where frame-based cameras may fail.
However, the asynchronous triggering property of neuromorphic sensors poses challenges for direct applications of these traditional computer vision approaches.

In terms of motion circuit modeling, SNNs are considered the most representative brain-inspired models for modeling human brain functions due to their biological plausibility~\cite{Guo2023fin, Roy2019nature}.
SNNs have shown excellent performance in the field of visual tasks~\cite{wu2018frontier}, including image classification. 
However, modeling the sequential velocity estimation problem for time-series learning poses two significant challenges for existing SNN models.
Firstly, the inherent sparsity, as well as low-precision spike representation, leads to a deficiency in numerical accuracy, potentially affecting their expressive ability in complex tasks.
Secondly, conventional SNN models, like Leaky Integrate-and-Fire (LIF) models, suffer from poor predictive capabilities. when confronted with the intricacies of time-series data.
Therefore, it is necessary to devise a novel neuron model and network structure to tackle these challenges effectively.

In this paper, we propose a brain-inspired framework, NeuroVE, specifically tailored to model vision motion circuits, enabling accurate estimations of the linear and angular velocities from event camera data.
To address the limitation of numerical precision, we design a novel ALIF neurons model inspired by astrocytes~\cite{Giantomasi2023SciRep}.
The ALIF model enhances the network's ability to express numerical information by adjusting the firing frequency of the SNNs, thus effectively mitigating the numerical inaccuracy caused by sparsity.
Besides, to address the limitation in time-series forecasting for SNNs, we propose an innovative ASLSTM module that integrates the ALIF model with long short-term memory. 
Further, we combine ALIF and ASLSTM to construct a brain-inspired velocity estimation framework.
To the best of our knowledge, this is the first brain-inspired approach to solving linear and angular velocity estimation problems by mimicking the brain's vision motion circuits.
We summarize our main contributions as follows:
\begin{itemize}
    \item We propose a brain-inspired framework that simulates the vision motion circuits of the brain for velocity estimation.
    The framework integrates an event camera for capturing motion information and employs brain-inspired SNNs to simulate spatial cells such as LS cells, AS cells, and time cells.
    \item Inspired by astrocytes, we design a brain-inspired neuron model, ALIF, that incorporates the neurotransmitter diffusion mechanism into the neuron dynamics.
    \item We introduce a brain-inspired SNN structure, ASLSTM, which integrates the dynamic property of ALIF neurons with the time-series forecasting capability of LSTM.
\end{itemize}

\section{Related Work}
\label{sec: related work}

\subsection{Numerical Regression with SNNs}
Spiking neural networks, a category of computational models inspired by biological neural systems, exhibit distinctive superiority in processing dynamic visual data.
The existing studies on numerical regression with SNNs usually employ population coding or membrane potential to encode numerical information.

Population coding represents values through the activation patterns of neuronal populations.
N. Iannella et al. use the inter-spike interval (ISI) coding approach to approximate a nonlinear function~\cite{Iannella2001NN}.
Lv et al. propose SNNs designed for time-series forecast tasks, which utilize fully connected (FC) layer coding to generate its outputs~\cite{lv2024icml}.
Employing the FC layer for output coding in SNNs represents an instantiation of population coding, which fundamentally uses the collective activity of neurons to encode information.
Gehrig et al. used SNN to estimate the angular velocity of the event camera~\cite{Gehrig2020icra}.
Their network architecture comprises six convolutional layers followed by a single fully connected layer.

The spike within SNNs are binary coding in nature.
The binary code typically implies a more extensive neuronal population to encode continuous values effectively.
To mitigate the redundancy in neuronal populations solely for numerical encoding, the approach of membrane potential encoding has been introduced.
The membrane potential encoding approach maps discrete spiking to continuous values by the non-spiking neurons.
Consequently, the absence of a reset mechanism in the neuron's membrane potential implies that the membrane potential can represent continuous values.

Recently, several applications have attempted to use membrane potentials to encode continuous values, including optical flow estimation and image generation, etc~\cite{Cuadrado2023Frontiers, Kamata2022AAAI}.
Furthermore, certain classification problems incorporate regression techniques to determine the probability distribution across various classes~\cite{Kim2020AAAI, Eshraghian2022Nano}.
Henkes et al. have provided an in-depth discussion on the regression challenges within SNNs and proposed a spiking long short-term memory (SLSTM) model to address numerical problems~\cite{Henkes2024RoyalScience}.

Nonetheless, despite these great progresses, the binary spike representation potentially limits the representational precision of existing models.
As information theory dictates, a higher precision representation of continuous values necessitates more information. 
Our approach efficiently encodes continuous values in SNNs through the introduction of a mechanism for the diffusion of membrane potentials.

\begin{figure*}[t]
    \centering
    \includegraphics[width=0.8\linewidth]{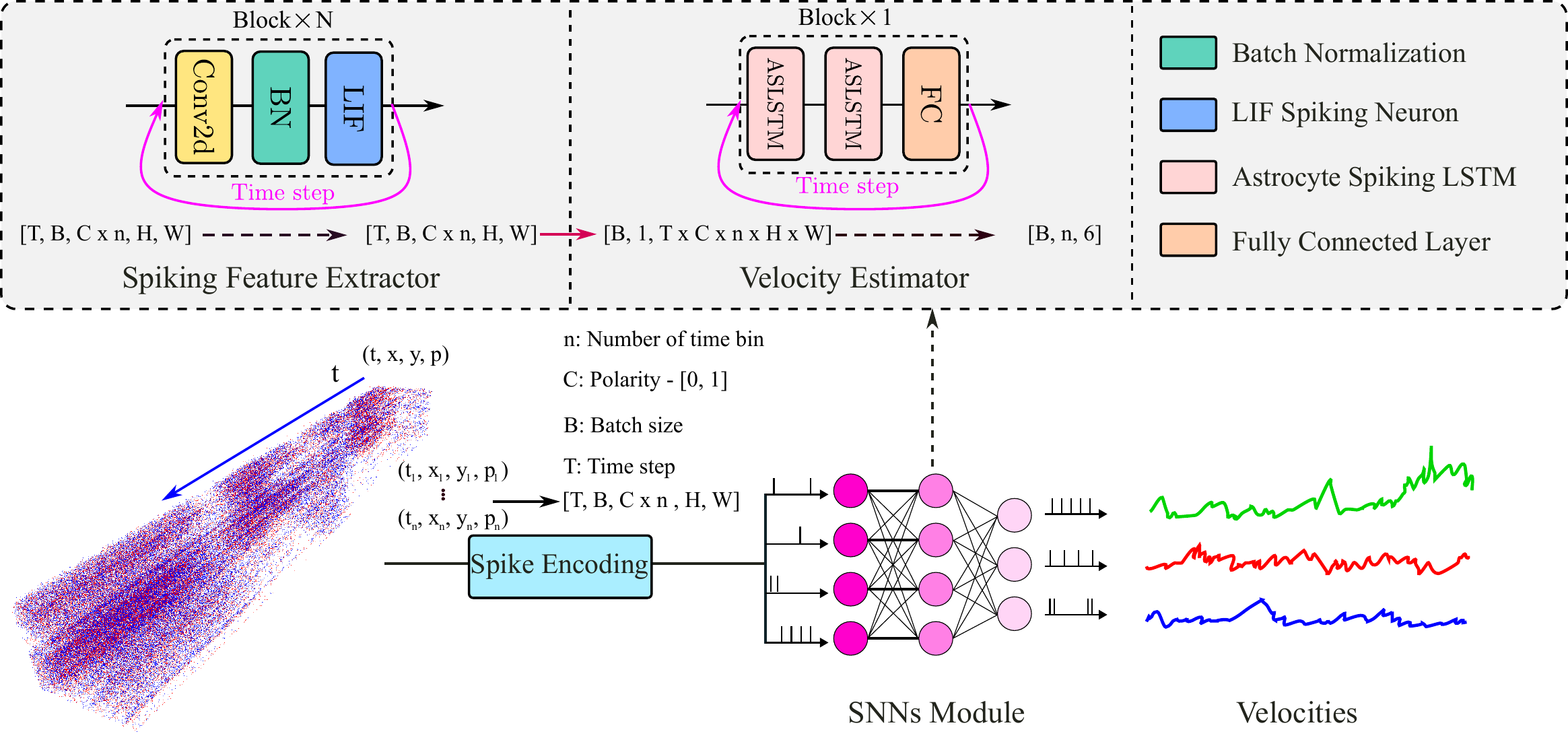}
    \caption{NeuroVE framework. The event data is denoted as $(t, x, y, p)$ and partitioned into $n$ chronological bins. These events are processed by spike coding and transformed into time-series spikes $[T, B, C \times n, H, W]$, where $T$ represents the time step, $B$ represents the batch size, $C$ represents the number of channels, $n$ represents the number of pieces into which the events are chronologically divided, $H$ and $W$ is the height and width, respectively. Finally, these time-series spikes are processed by the spiking feature extractor and the velocity estimator to infer the linear and angular velocities directly.}
    \label{fig: network}
\end{figure*}
\subsection{Event-based Velocity Estimation}

While significant progress has been achieved in navigation and localization, exploring first-order kinematics remains underexplored.
In the current event-based odometry approach, some studies have optimized the pose and velocity of the event camera concurrently as state variables, which is especially prevalent in visual-inertial odometry (VIO) processes.
Mueggler et al. have introduced a continuous-time VIO framework. 
Unlike discrete-time VIO, which estimates pose and velocity separately and may be susceptible to inconsistencies, the continuous-time approach offers a coherent and unified representation of pose and velocity~\cite{Mueggler18tro}.
In~\cite{Bryner19icra}, a method is proposed for simultaneously estimating an event camera's velocity and pose, leveraging a known photometric 3D map.
They derived an intensity-change residual loss and determined the pose and velocity of the event camera by nonlinear optimization.

Another vein of research focuses on the unique asynchronous spatio-temporal properties of event cameras for the direct estimation of velocity.
A pioneering work goes back to the contrast maximization (CM) framework introduced by Guillermo et al~\cite{Gallego18cvpr}.
However, the computational cost associated with CM often impedes running in real-time.
Peng et al. have introduced a geometry-based velocity estimation method. They have formulated a closed-form solution for linear velocity estimation by employing the trifocal tensor~\cite{Peng2021BMVC}.
Li et al. have introduced a calibration method established on linear velocity correlation~\cite{lx2024ral}. 
They estimate the camera's linear velocity under the constraint of linear motion for calibration.
This geometry-based methodology assumes robust feature extraction to capture motion information effectively.
More recently, the contributions of Lu et al. are particularly noteworthy, which proposed an advanced velometer~\cite{lu2024rss}.
They employed stereos event cameras to determine the depth and normal flow and integrated IMU data within a continuous-time framework.

The majority of existing approaches rely on traditional computer vision pipelines, which fail to fully harness the potential of event cameras.
Our brain-inspired computational paradigm effectively integrates the asynchronous spatio-temporal characteristics of event camera data.

\section{Methodology}
\label{sec:method}

\subsection{NeuroVE Framework}

The NeuroVE framework is depicted in Fig.~\ref{fig: network}, which consists of three main components: spike encoding, spiking feature extractor, and velocity estimator. 
\begin{itemize}
    \item \textbf{Spike Encoding}: 
    The data pre-processing phase, which categorizes event data over a specified time interval based on its polarity, assigns events with negative polarity to channel 0 and positive polarity to channel 1 in each chronological bin.
    \item \textbf{Spiking Feature Extractor}: 
    The spiking feature extractor comprises $N$ sequential blocks, each equipped with a convolutional layer, a batch normalization layer~\cite{Zheng2021aaai}, and a LIF neuron.
    \item \textbf{Velocity Estimator}: 
    The velocity estimator receives the high-level features generated by the Spiking Feature Extractor. 
    Furthermore, the velocity estimator integrates ALIF neurons with ALSTM, enabling accurate estimation.

\end{itemize}

Activation of LS and AS cells in the brain has been demonstrated to exhibit a nonlinear correlation with the linear and angular velocities~\cite{Spalla2022NC}.
Moreover, time cells do not encode absolute temporal information, such as circadian rhythms. 
Instead, they are specialized in encoding sequential information~\cite{Eichenbaum2014nrn}.

Accordingly, the operational mechanism of these cells can be effectively modeled using SNNs.
In the spike encoding phase, we simulate the function of time cells by mapping sequential event camera data into $n$ chronological bins.
Besides, we regard the SNNs as a nonlinear mapping, like LS and AS cells, that transforms high-dimensional event data into a six-dimensional velocity vector.
Once trained, the network predicts velocity at every moment, even with novel scenarios, by the inherent temporal structure of event data. 
This time-series forecasting-based velocity estimation approach enables our model to adapt to known data distributions and generalize to new environments.

In the following subsections, we dive into the foundational design principles and implementation details of the ALIF (Sec.~\ref{subsec: alif}) and ASLSTM (Sec.~\ref{subsec: aslstm}) models.

\subsection{Astrocytes Leaky Integrate-and-Fire Neurons}
\label{subsec: alif}
In this section, we introduce the ALIF neuron model, which can effectively solve the instability of SNNs in numerical regression problems.

The basic LIF model can be written as
\begin{equation}
    \begin{aligned}
    \tau \frac{dV(t)}{dt} &= -(V(t)-V_{rest}) + RI(t),\\
    V(t) &= V_{rest}, ~if~ V(t)>V_{th},
    \end{aligned}
 \label{eq: lif model}
\end{equation}
where $V(t)$ is the membrane potentials at time $t$, $V_{rest}$ is the neuronal resting potential, $I(t)$ is the pre-synaptic input current, $R$ is 
the neuronal resistance and $\tau$ is the leakage constant.
When the membrane potential exceeds the threshold $V_{th}$, the neuron fires an action potential, followed by a membrane potential returns to the resting potential $V_{rest}$.
If the membrane potential fails to reach the threshold $V_{th}$, it undergoes leakage at $1 / \tau$ rate.

However, the sparsity firing pattern induced by the LIF neuron model poses challenges for numerical regression applications.
Sparse signals cannot inherently maintain high precision values with integrity.
The neuron's membrane potential is initialized at the resting level, requiring sufficient time to accumulate charge to reach the threshold for firing.
While increasing the time step relieves this issue to a degree.
It concurrently leads to exponentially increasing network dimensions and memory usage.
Moreover, the numerical challenges introduced by sparsity are not effectively addressed by simply increasing the time step, particularly in the time series forecasting problem.

Astrocytes orchestrate the synchronization of neural signals through the modulatory control of neurotransmitter transmission among neurons~\cite{Giantomasi2023SciRep}.
Inspired by astrocytes, we introduce a diffusion model grounded in Fick's law~\cite{Wheatley1998Biosystems}, designed to simulate the neurotransmitter transfer from astrocytes to neurons.
\begin{equation}
    J(i,j) = D \frac{\partial C}{\partial t} = D(V_i-V_j),
    \label{eq:fick}
\end{equation}
where $J(i,j)$ represents the diffusion equation between the $i$th and $j$th neurons, $D$ is the diffusion coefficient, and $C$ represents the concentration of neurotransmitters on the postsynaptic terminal.
However, to avoid the introduction of superfluous hyperparameters, the $C$ is defined as the differential in membrane potentials across neurons. 
Consequently, we obtain the ALIF model.
\begin{equation}
\begin{aligned}
    \tau \frac{dV(t)}{dt} &= -(V(t)-V_{rest}) + R(I(t) + J),\\
    V(t) &= V_{rest}, ~if~ V(t)>V_{th}.
    \label{eq: alif model}
\end{aligned}
\end{equation}

Eq.~\ref{eq: alif model} implies that the membrane potential is transmitted to adjacent neurons. To derive a gradient-friendly solution for Eq.~\ref{eq: alif model}, it is effective to adapt the formula for iterative computation and assume that $V_{rest}$ is equal to 0.
We employ the Euler method for the first-order differential equation~\cite{wu2018frontier} as
\begin{equation}
\begin{aligned}
    V^{t+1}_j &= (1-\frac{dt}{\tau})V^t_j + \frac{dt}{\tau}R(I^t_j + J(i,j)),\\
    V^t_j &= V_{rest}, ~if~ V^t_j>V_{th},
    \label{eq: euler alif model}
\end{aligned}
\end{equation}
where the upper index $t$ and lower index $j$ denote time moment and neuron index, respectively. The pre-synaptic current $I(t)$ is the neuron's input signal $X$, and $R$ is the operator $f(\cdot)$, encompassing various forms such as convolutional or linear operators.
Additionally, we denote $(1-\frac{dt}{\tau})$ as $\alpha$ and $\frac{dt}{\tau}$ as $\beta$.
Thus, Eq.~\ref{eq: euler alif model} can be simplified to
\begin{equation}
\begin{aligned}
    V^{t+1}_j &= \alpha V^t_j + \beta(f(X_j + D(V_i^{t_f}-V^{t_0}_j))),\\
    V^t_j &= 0, ~if~ V^t_j>V_{th},
    \label{eq: euler alif model2}
\end{aligned}
\end{equation}
where $V_j^{t_0}$ is the initial membrane potential of the $j$th neuron, $V_i^{t_f}$ is the membrane potential of the $i$th neuron at $t_f$ moment.

Finally, by simplifying the constant term in Eq.~\ref{eq: euler alif model2} and assuming $V^{t_0}_j=0$, we derive an iterative formulation for the ALIF neuron model.

\begin{subequations}
\begin{align}
    V^{t+1}_j &= (1-S_j^t)V^t_j+f(X_j+DV^{t_f}_i),
    \label{subeq: iter alif a}
    \\
    S_j^t &= \mathcal{H}(V^t_j),
    \label{subeq: iter alif b}
    \\
    \mathcal{H} &= \left\{
    \begin{aligned}
    0, \quad V^t_j < V_{th}\\
    1, \quad V^t_j \geq V_{th}\\
    \end{aligned}
    \right
    ..
    \label{subeq: iter alif c}
\end{align}
\label{eq: iter alif}
\end{subequations}

The iterative process of the ALIF neuron model involves two sequential phases: updating the neuron's state (Eq.~\ref{subeq: iter alif a}) and spike firing (Eq.~\ref{subeq: iter alif b}).
When the membrane potential exceeds the threshold value, the neuron generates a spike.
Subsequently, at the next time step, the neuron will reset its membrane potential depending on the occurrence of spikes. 
Furthermore, due to the non-differentiable property of Eq.~\ref{subeq: iter alif c}, alternative surrogate function~\cite{wu2018frontier} is employed during the training phase.

\begin{figure}[t]
    \centering
    \includegraphics[width=1\linewidth]{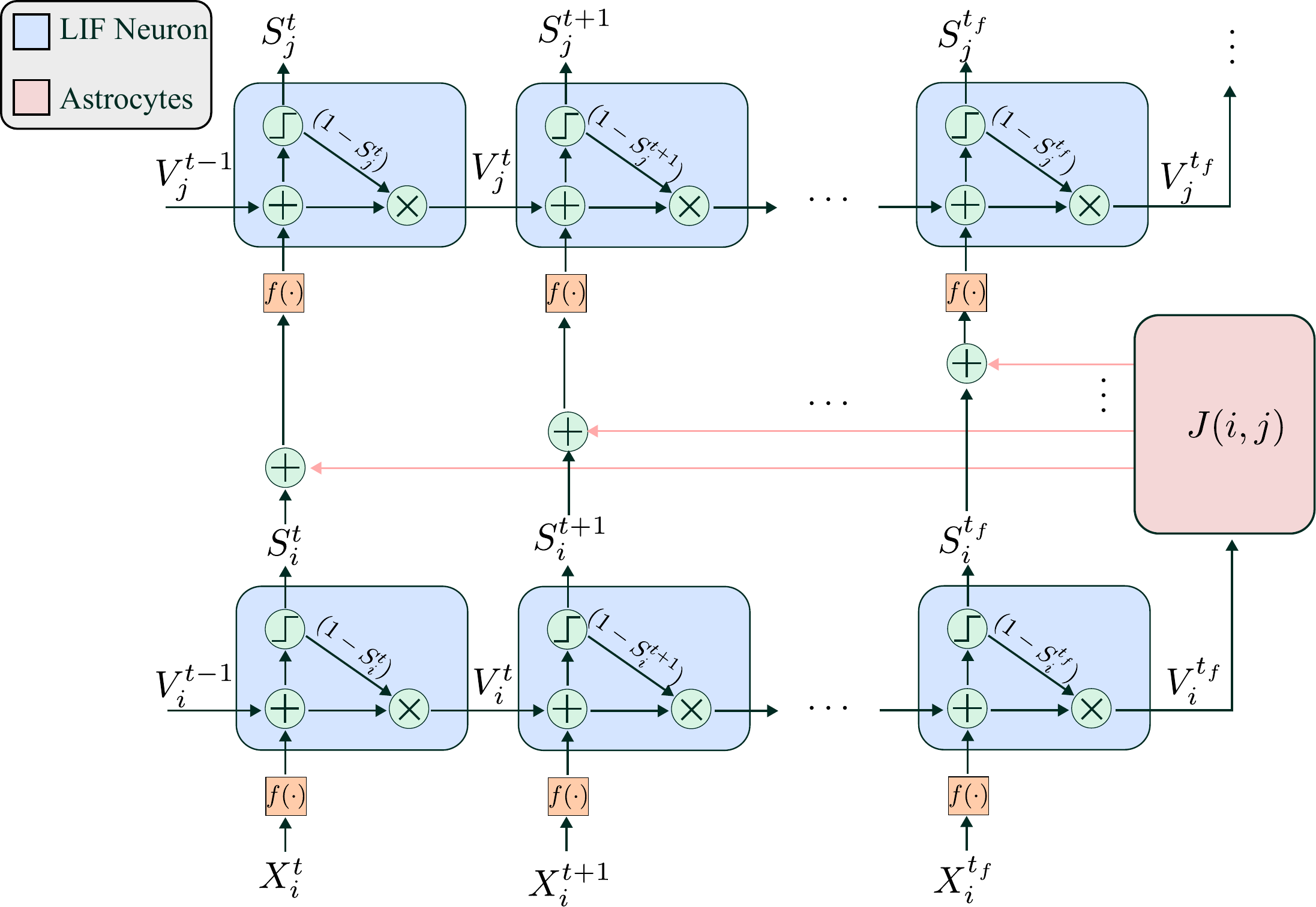}
    \caption{The illustration describes the diffusion mechanism of membrane potentials in the two-dimensional computational graph of SNNs. Here, the blue block represents the LIF neurons, and the pink block represents the astrocyte-inspired diffusion mechanism. }
    \vspace{-0.5cm}
    \label{fig: alif}
\end{figure}

Fig.~\ref{fig: alif} illustrates the phenomenon where the membrane potential of the $i$-th neuron diffuses into the $j$-th neuron at a predefined rate.
In SNNs, inputs propagate concurrently across both temporal and spatial dimensions, which thereby can be unfolded as a two-dimensional computational graph.

We employ a temporally prioritized sequential computational method for our neural network.
Computing the activation state of each neuron at a given time step before moving on to the next. 
This sequential computation ensures that astrocytes are informed of the membrane potential at the final step of the preceding neuron and enables the effective diffusion of these potentials across the network.

\subsection{Astrocytes Spiking Long Short-term Memory Network}
\label{subsec: aslstm}
The task of linear-angular velocity regression for an event camera is formulated as a time-series model, capturing the sequential progression of velocities over time.
Despite the spatio-temporal dynamics inherent to SNNs, our empirical findings indicate that directly employing plain SNNs into a time-series model is prone to encountering gradient explosion issues during the backpropagation phase.

Accordingly, we first introduce the ASLSTM for the time-series forecasting task, built upon the ALIF model introduced in \ref{subsec: alif}.
The ASLSTM introduces supplementary gating mechanisms, similar to standard LSTM~\cite{Hochreiter1997LSTM} and SLSTM~\cite{Henkes2024RoyalScience}, enhancing the network's spatial-temporal dynamics.

The ASLSTM can be formulated as
\begin{equation}
\begin{aligned}
    c^t &= \sigma(f^t) \otimes c^{t-1} + \sigma(i^t) \otimes tanh(g^t), \\
    h^t &= \sigma(o_t) \otimes tanh(c_t), \\
    v^t &= \mathcal{A}(h^t, x^t, v^{t-1}),\\
    \zeta &= \mathcal{O}(v^{t_f}, x^{t_f}),\\
    s^t &= \mathcal{H}(v^t), \\
    \mathrm{with} \\
    i^t &= W_ix^t + U_iv^{t-1}, \\
    f^t &= W_fx^t + U_fv^{t-1}, \\
    g^t &= W_gx^t + U_gv^{t-1}, \\
    o^t &= W_ox^t + U_ov^{t-1},
\end{aligned}
\label{eq: aslstm}
\end{equation}
where $i^t$, $o^t$, $f^t$, and $g^t$ are inputs of the input gate, output gate, forget gate, and cell state, respectively. 
The signals are consistent with the standard LSTM, where $\sigma$ denotes the sigmoid function, $tanh$ denotes the tangent hyperbolic function, and $\otimes$ is the Hadamard product.
$\mathcal{H}$ is the Heaviside step function formulated in Eq.~\ref{subeq: iter alif c}.
$\mathcal{O}(\cdot)$ is the output neurons, which can be formulated by 
\begin{equation}
    \mathcal{O}(v^{t_f}, x^{t_f}) = \kappa v^{t_f}+f(x^{t_f} + J(t-1,t)), \quad 0 < \kappa < 1.
\end{equation}

$\mathcal{A}(\cdot)$ is the state update function of the ALIF neuron, which can be written as
\begin{equation}
    \mathcal{A}(h^t, x^t, v^{t-1}) = \alpha(1-s^{t-1})h^t + f(x^t + J(t-1,t)),
    \label{eq: su of alif}
\end{equation}
$J(t-1,t)$ represents the diffusion across various time steps within the same neuron.

\begin{figure}[t]
    \centering
    \includegraphics[width=1\linewidth]{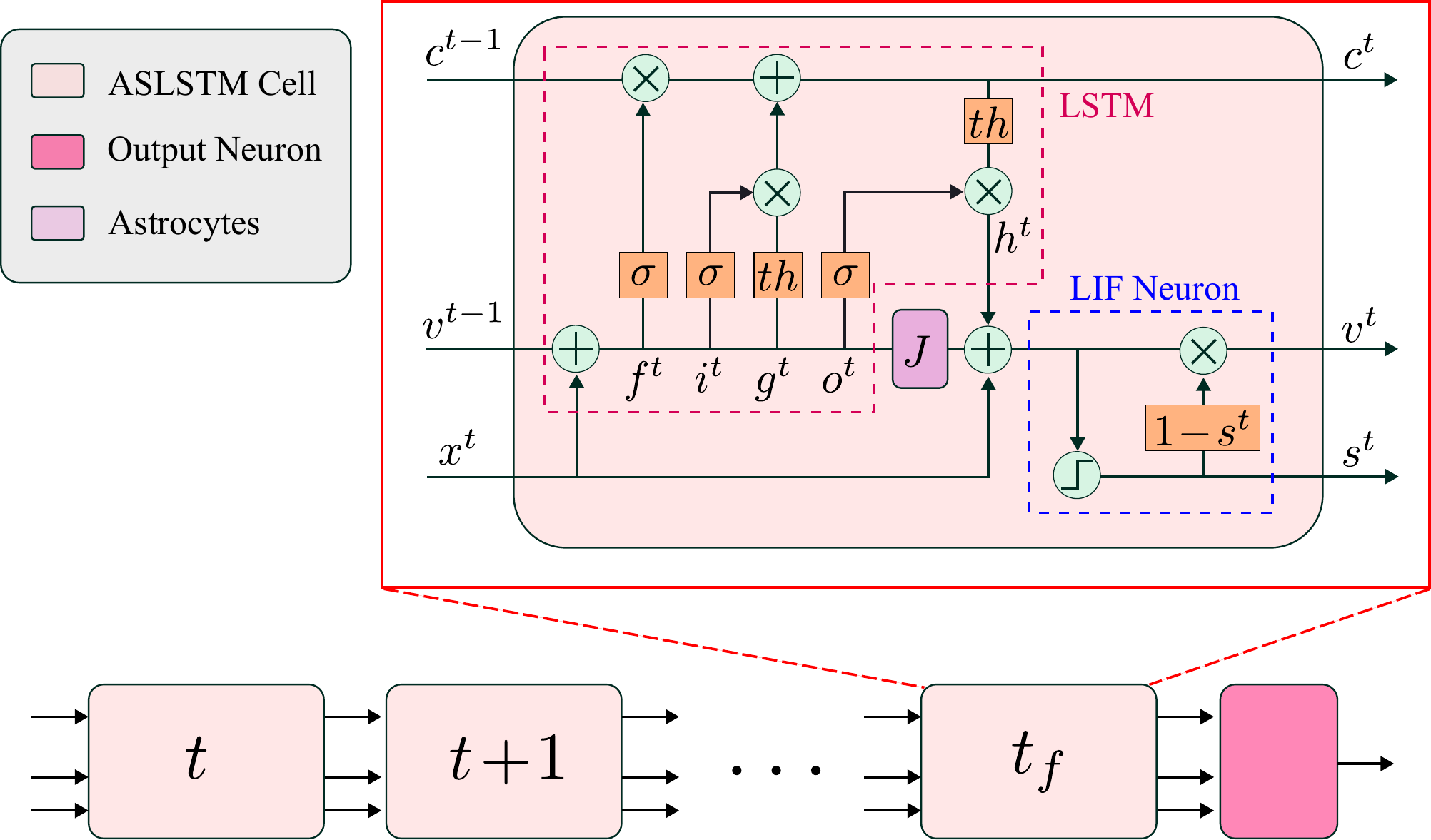}
    \caption{Elaborate on the mechanism by which astrocytes diffuse membrane potentials and illustrate the integration process with the LIF neurons within the ASLSTM.}
    \vspace{-0.5cm}
    \label{fig: aslstm}
\end{figure}

Fig.~\ref{fig: aslstm} represents the internal architecture and synaptic connections within the ASLSTM.
The inputs of ASLSTM propagate exclusively along the temporal dimension, rather than spatially, at each time step.
Upon reaching the final time step, the output neuron is engaged to emit membrane potentials that represent the continuous values.
The region enclosed by the red dashed line in Fig.~\ref{fig: aslstm} is the standard LSTM cell.
Due to the presence of LIF neurons within the ASLSTM cell, the inputs to ASLSTM comprise three parts: long-term memory $c^{t-1}$,  membrane potential $v^{t-1}$, and spikes $s^{t-1}$.

\subsection{Training Detials}
The loss $L$ is constituted by two distinct components: the angular velocity loss $L_a$ and the linear velocity loss $L_l$:
\begin{equation}
\begin{aligned}
    L &= L_{a} + L_{l} \\
    L_i &= \frac{1}{2} \sqrt{e_i^Te_i}, \quad i \in \{a, l\}
\end{aligned}
    \label{eq: loss}
\end{equation}
where $e_{a} = \omega_{gt} - \omega$, $e_{l} = l_{gt} - l$, $\omega$ is the angular velocity and $l$ linear velocity, respectively.
Both angular and linear velocities are defined within the three-dimensional Euclidean space $\mathbb{R}^3$.
Since there is a substantial numerical value disparity between angular and linear velocities. 
The gradients associated with them vary by several orders of magnitude throughout the training process.
Consequently, we implemented a dynamic scaling of the loss functions for angular and linear velocities to mitigate this discrepancy.
The dynamic scaling balanced the influence of angular and linear velocities within the loss function.

As shown in Fig.~\ref{fig: network}, our training process is multi-step.
Firstly, the event data is partitioned into $n$ chronological bins. 
These events are assigned to different channels according to their polarity.
Secondly, the polarity-divided events are subsequently converted into a time series spike tensor $[T, B, C \times n, H, W]$.
Finally, these time-series spikes are fed into the SNN module to estimate the velocities $[B, n, 6]$, where the angular velocities are represented by Euler angles.

In the computational process, we sequentially allow each layer to complete its computations at the time step. 
The resultant information is passed on to the subsequent layer only after each layer has fully processed the data at a given time step.
The SNNs are trained by the Adam optimizer within the public PyTorch framework~\cite{Paszke17nipsw}.

\section{Experiments}
\label{sec:evaluation}
    
In the experimental section, we designed three experiments to evaluate our NeuroVE framework. 
Initially, we evaluated the performance of the ALIF neurons and the ASLSTM in nonlinear regression and time-series forecasting experiments.
Subsequently, we evaluated the performance of NeuroVE in the linear-angular velocity estimation experiments. 
Finally, real-world experiments confirmed the effectiveness of NeuroVE.

\subsection{Experimental Configuration}

In this section, we detail the experimental configurations for the three experiments and the metrics employed to evaluate the results.

\subsubsection{Nonlinear regression and time-series forecasting experiments}
The \textit{nonlinear regression and time-series forecasting experiments} aim to assess the ASLSTM's ability to perform numerical precision and time-series forecasting.
The existing studies on SNNs for nonlinear numerical regression experiments frequently employed simple configurations.
In addition, the proficiency of SNNs in time-series prediction is inferior to ANNs.
To this end, we have designed an experiment to fit the $y=sin(x)$ curve, encompassing numerical regression and predictive components.

We randomly generated 100 instances of $sin(x+d_i)$, each incorporating phase offsets $d_i$, allocating 97 sequences for training and reserving 3 sequences for validation.
Furthermore, the whole training dataset is comprised of 1000 time steps.
The experiment fits the data across the initial 0 to 1000 time steps and forecasts the outcomes for the subsequent 1000 to 2000 time steps.

Furthermore, we undertook a comparative analysis of the firing frequencies across various neuron types when subjected to identical random signals to demonstrate the dense firing frequency feature of ALIF neurons. 

\subsubsection{Linear-angular velocity estimation experiments}
The \textit{linear-angular velocity estimation experiments} is specifically designed to evaluate the framework's performance in accurately determining velocities from event camera data.
This evaluation was performed on two distinct datasets: the MVSEC dataset\cite{Zhu18ral} and a synthetic dataset generated by the CARLA simulator\cite{Dosovitskiy17corl}, which provides a controlled environment for testing.
For the training phase, we used the $ out\_door\_day2$ dataset and a subset of the $ out\_door\_day1$ dataset.
Following the training phase, we evaluated the model's performance in an unused portion of the $ out\_door\_day1$ dataset.

\subsubsection{Real-world experiments}
In real-world experiments, our objective is to confirm the efficacy of our proposed NeuroVE framework on the robot. 
To this end, we affixed an event camera to a four-wheeled robot and tracked its movement along a corridor within a room (see Fig.~\ref{fig: real-world}).
We recorded a total of four distinct trajectories during the experiments to demonstrate the performance of NeuroVE. 
The outputs from the multi-sensor fusion algorithm were used as the ground truth for velocity estimation.
To guarantee the precision of our evaluation, we systematically sampled a subset of each trajectory for a validation set, deliberately excluded from the training phase.

\subsubsection{Metrics}
In order to comprehensively evaluate the performance of the proposed NeuroVE framework, we employ different metrics to quantify the performance in the experiments.
We have selected the Root Mean Square Error (RMSE) as our primary metric to verify the model's numerical accuracy and time-series prediction performance. 
Besides, we have also introduced the Relative Error (RE) as a complementary metric to RMSE as the primary evaluation metric. 
As a dimensionless measure, RE provides a clear and direct insight into the magnitude of the error as a percentage of the actual value. 
The formulas are shown as follows,
\begin{equation}
\begin{aligned}
    RMSE(x) &= \sqrt{\frac{1}{n} \sum_i^n \Vert x_{gt}^i - x^i \Vert^2}, \\
    RE(x) &= \frac{1}{n} \sum_i^n \frac{ \Vert x_{gt}^i - x^i \Vert}{x_{gt}^i},
\end{aligned}
    \label{eq: metric}
\end{equation}
where $x^i_{gt}$ is the $i$th ground truth, $x^i$ is the $i$th prediction value.

\subsection{Results of Nonlinear Regression and Time-series Forecast Experiments}
In the experiments, we evaluate our method and compare its performance with existing pipelines listed as follows:
\begin{itemize}
    \item ASLSTM (Ours): The method proposed in Sec.~\ref{subsec: aslstm}, and the network streamlined architectural design, comprising two layers of ASLSTM cells.
    \item ASLSTM-w/o: The proposed networks utilize LIF neurons~\cite{wu2018frontier} in place of ALIF neurons, thereby excluding the astrocyte-inspired diffusion mechanism.
    \item LTC~\cite{Yin2023NMI}: The SNNs incorporate liquid time-constant spiking neurons designed to address the challenges associated with long-term time series prediction.
    \item SLSTM~\cite{Henkes2024RoyalScience}: The SLSTM proposed for regression using spiking neural networks.
\end{itemize}

\begin{figure}[t]
    \centering
    \includegraphics[width=1\linewidth]{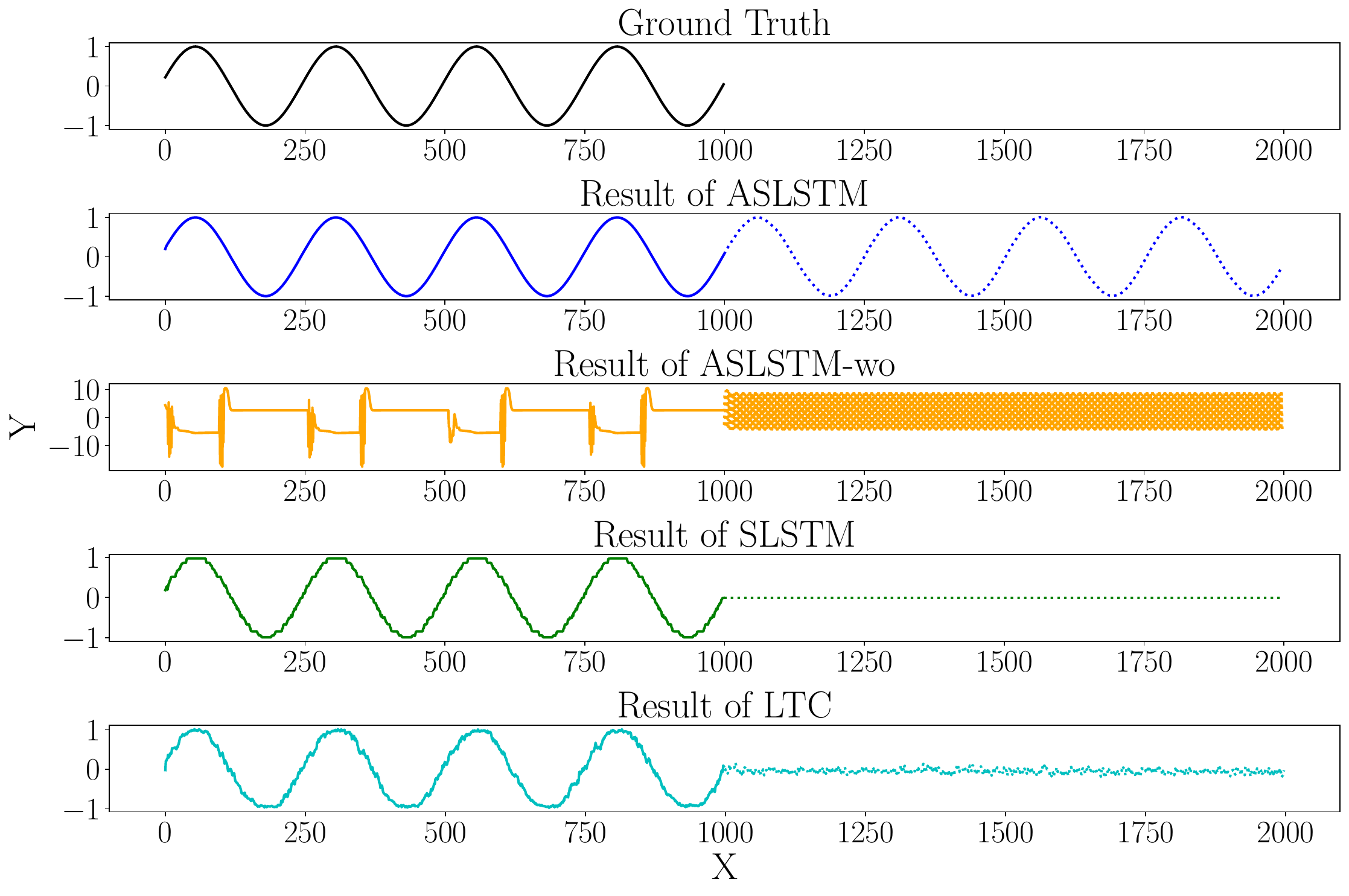}
    \caption{The graph of the results of numeric regression and time-series forecasting for $y = sin(x)$.}
    \label{fig: regression result}
\end{figure}
\begin{figure}[t!]
  \centering
  \subfigure[t][\small
  {Randomly generated current signals.}]{
  \includegraphics[width=1\columnwidth]{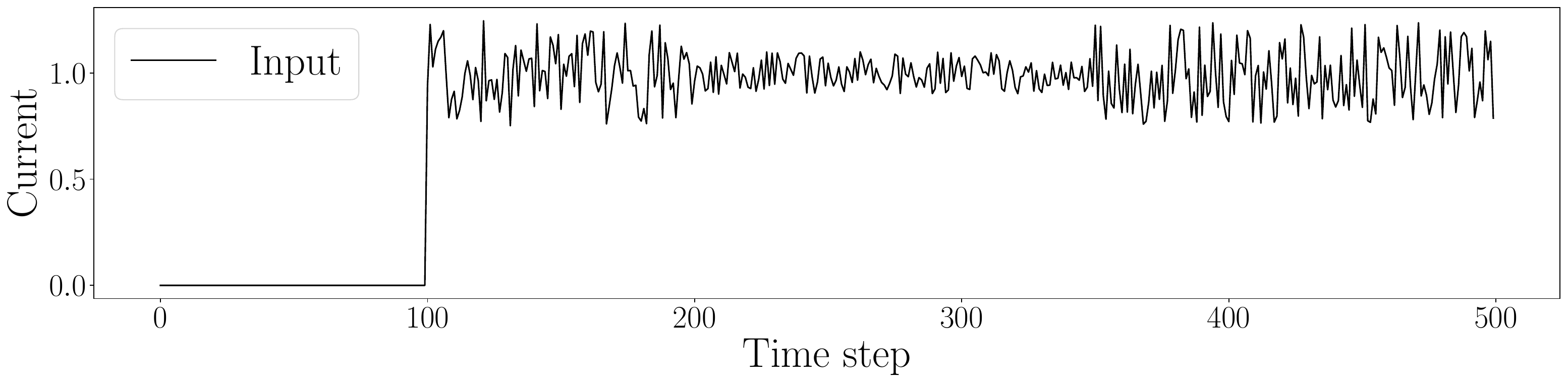}
  \label{subfig: nueron-spike input}}\!\!
  
  \subfigure[t][\small{Response of ALIF neuron.}]{
  \includegraphics[width=1\columnwidth]{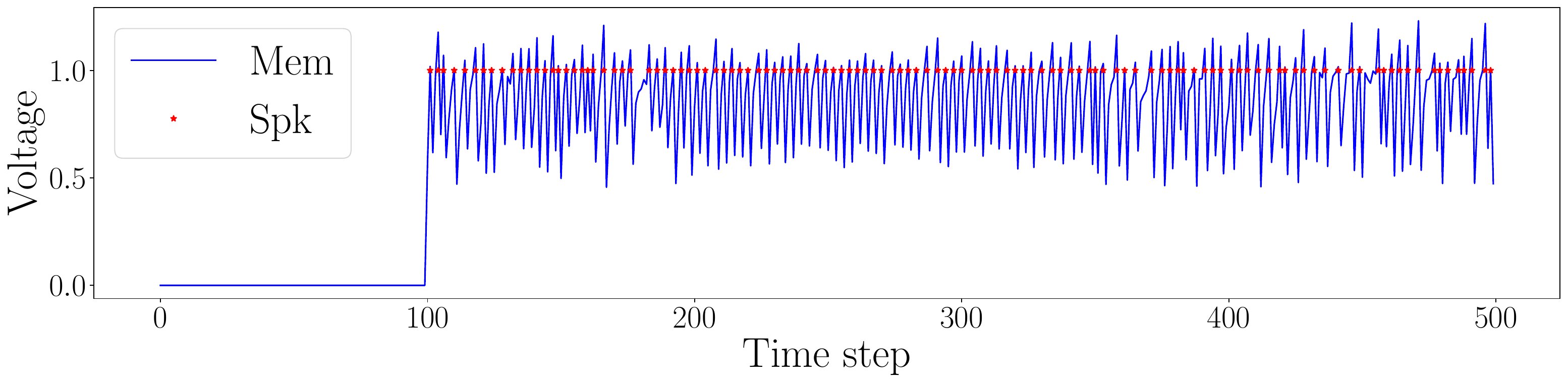}
  \label{subfig: nueron-spike alif}}\!\!
  
  \subfigure[t][\small{Response of LIF neuron.}]{
  \includegraphics[width=1\columnwidth]{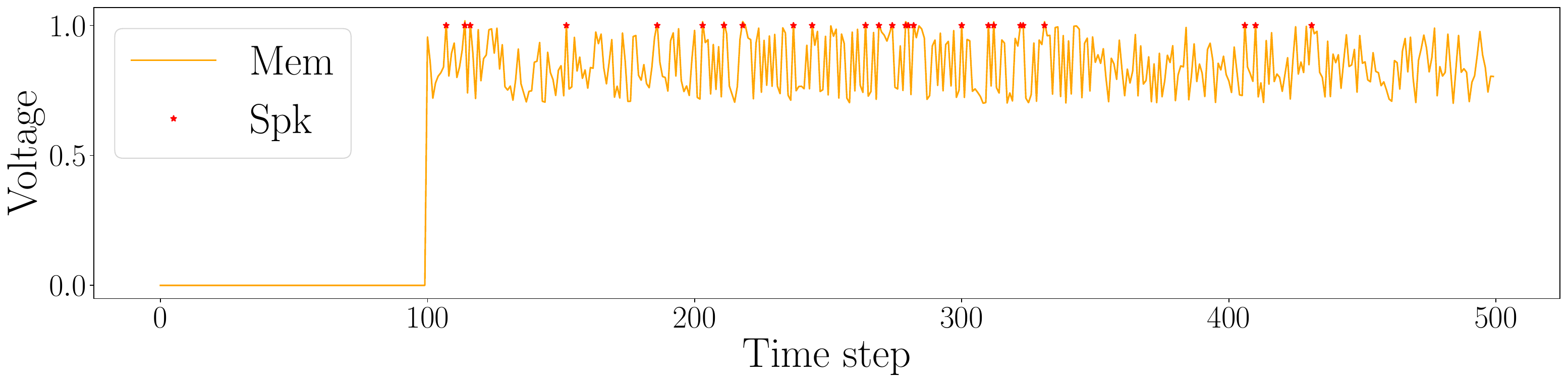}
  \label{subfig: nueron-spike lif}}\!\!

  \subfigure[t][\small{Response of LTC neuron.}]{
  \includegraphics[width=1\columnwidth]{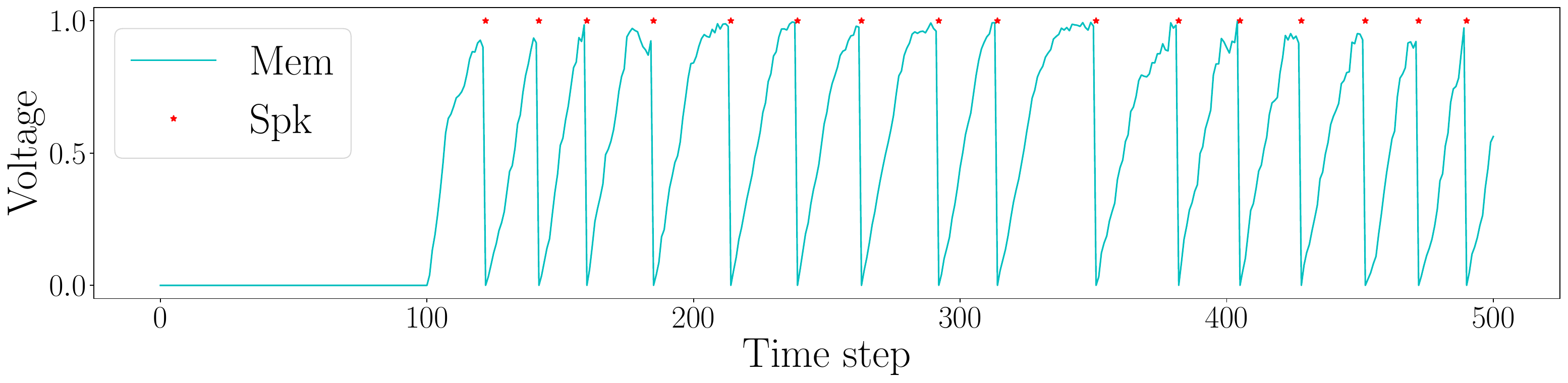}
  \label{subfig: nueron-spike ltc}}\!\!
  
  \caption{Results of neuronal firing frequency.}
  \label{fig: neuron spike}
\end{figure}

 \begin{table}[t]
     \caption{RMSE$^{\dag}$ for numerical regression tasks.}%
     \centering%
     \begin{threeparttable}
     \begin{tabular}{ccccc}%
         \toprule%
         Sequence   & SLSTM & LTC & ASLSTM-w/o & ASLSTM(Ours)   \\
         \midrule%
         1 &  0.86 & 3.16  & failed & \textbf{0.31} \\
         2 &  0.84 & 3.67  & failed & \textbf{0.30} \\
         3 &  0.87 & 3.92  & failed & \textbf{0.31} \\
         \bottomrule%
     \end{tabular}
     \begin{tablenotes}
         \footnotesize
         \item[1] RMSE$^{\dag}$ = RMSE $\times$ $10^3$.
     \end{tablenotes}
     \end{threeparttable}
     \label{tab: regression task}
     \vspace{-2.0cm}
\end{table}

 \begin{table}[t]
     \caption{RMSE for time-series forecast tasks.}%
     \centering%
     \begin{tabular}{cccccc}%
         \toprule%
         Sequence  & SLSTM &  LTC & ASLSTM-w/o & ASLSTM(Ours)   \\
         \midrule%
         1 & 0.50 & 0.53  & failed & \textbf{0.02} \\
         2 & 0.88 & 0.86  & failed & \textbf{0.07} \\
         3 & 0.95 & 1.04  & failed & \textbf{0.11} \\
         \bottomrule%
     \end{tabular}
     \label{tab: time series task}
     \vspace{-0.8cm}
\end{table}

We begin by discussing the numerical experiment part, depicting the 0-1000 time steps in Fig.~\ref{fig: regression result}.
The numerical experiment part shows the smoothest curve of our results.
In contrast to other methods, they exhibit fluctuations due to SNN numerical instability. 
Tab~\ref{tab: regression task} further illustrates the significant improvement of our method, with an RMSE$^{\dag}$ of approximately 0.3, markedly better than the 0.8 of SLSTM and 3 of LTC.
Since the $sin(x)$ oscillates between an amplitude range of -1 to 1, the resulting RMSE is inherently small. 
To facilitate a clear comparison, we introduced RMSE$^{\dag}$, which magnified 1000 times from RMSE.

In the time-series forecasting part, depicted in the 1000-2000 time steps of Fig.~\ref{fig: regression result}. 
The experimental results indicate that most alternative methods struggle with time-series forecasting during 1000-2000 time steps. 
Tab.~\ref{tab: time series task} quantitatively confirms our method's superior performance in time-series forecasting. 
Furthermore, despite acceptable RMSE values for other methods, Fig.~\ref{fig: regression result} exposes their deficiencies in time-series forecasting capabilities.

Finally, Fig.~\ref{fig: neuron spike} indicates that our method improves SNN's numerical stability through increased firing frequency.

\subsection{Results of Linear-Angular Velocity Estimating Experiments}
In this section, our primary objective is to validate the efficacy of brain-inspired methodologies in the estimation of velocities and angular velocities from event cameras. 
We have selected the MVSEC dataset, which offers accurate ground truth of trajectory.
By differentiating these poses, we obtain the ground truth of linear and angular velocities, which provide a benchmark for model evaluation.
Our methods and baseline methods are listed as follows:
\begin{itemize}
    \item NVE (Ours): NeuroVE is proposed in Sec.~\ref{sec:method}.
    \item NVE-w/o: The method employs LIF neurons~\cite{wu2018frontier} instead of ALIF neurons.
    \item SNN-ANG~\cite{Gehrig2020icra}: The approach uses SNN to implement angular velocity regression.
\end{itemize}

During the training phase, we do not utilize all the trajectories for model training. 
Instead, we allocate a portion for training and reserving another for the validation and testing phases.
Ultimately, we employ the RMSE and RE as the primary evaluation metric.

 \begin{table}[t]
     \caption{Evaluation on MVSEC dataset }%
     \centering%
     \renewcommand{\arraystretch}{1.5}
     \begin{threeparttable}
     \begin{tabular}{cccccccc}%
         \toprule%
         Sequence & Error  & \multicolumn{2}{c}{SNN-ANG} & \multicolumn{2}{c}{NVE-w/o} & \multicolumn{2}{c}{NVE(Ours)}   \\
        -  &  -   &  $v_a$&$v_l$ & $v_a$&$v_l$ & $v_a$&$v_l$ \\
         \midrule%
        
         \multirow{2}{*}{out\_door\_day1} 
         & RMSE$^*$ & 0.91&0.7 & 0.64&0.3 & \textbf{0.11}&\textbf{0.12} \\
         
         & RE   & 1.63&0.45 & 0.89&0.22 & \textbf{0.15}&\textbf{0.10} \\
         \bottomrule%
     \end{tabular}
     \begin{tablenotes}
         \footnotesize
         \item[1] RMSE$^*$: RMSE of angular velocity ($v_a$) magnified by a factor of 100.
         \item[2] The unit of linear velocity ($v_l$) is $m/s$ and angular velocity ($v_a$) is $deg/s$.
     \end{tablenotes}
     \end{threeparttable}
     \label{tab: velocity task}
\end{table}

 \begin{table}[t]
     \caption{Evaluation on synthetic dataset}%
     \centering%
     \renewcommand{\arraystretch}{1.5}
     \begin{tabular}{cccccccc}%
         \toprule%
         Sequence & Error  & \multicolumn{2}{c}{SNN-ANG} & \multicolumn{2}{c}{NVE-w/o} & \multicolumn{2}{c}{NVE(Ours)}   \\
        -  &  -   &  $v_a$&$v_l$ & $v_a$&$v_l$ & $v_a$&$v_l$ \\
         \midrule%
        
         \multirow{2}{*}{Seq 1} 
         & RMSE$^*$ & 5.77&0.75 & 5.37&0.30 & \textbf{3.56}&\textbf{0.28} \\
         
         & RE   & 1.84&0.91 & 1.02&0.34 & \textbf{0.73}&\textbf{0.32} \\
         \cline{2-8}
         \multirow{2}{*}{Seq 2} 
         & RMSE$^*$ & 4.41&0.99 & 3.82&0.38 & \textbf{3.37}&\textbf{0.35} \\
         & RE   & 1.92&0.92 & 1.14&0.38 & \textbf{1.01}&\textbf{0.31} \\
         \bottomrule%
     \end{tabular}
     \label{tab: velocity task on carla}
     \vspace{-1.0cm}
\end{table}

The experimental results are presented in Tab.~\ref{tab: velocity task} and Tab.~\ref{tab: velocity task on carla}.
Our method shows better performance in estimating linear and angular velocities, outperforming other advanced approaches.
In Tab.~\ref{tab: velocity task}, our method demonstrates superior performance over other approaches, achieving an RMSE$^*$ of 0.11 for angular velocity and 0.12 for linear velocity.
The RMSEs of angular and linear velocities are not balanced.
Most of the samples in the dataset exhibit linear motion, resulting in very small values of mean angular velocity.
The RMSE$^*$ represents the RMSE of angular velocity by a factor of 100.

In addition, we conducted numerous randomized experiments within the MVSEC dataset, which yielded results indicating that NeuroVE realizes a performance improvement of approximately 60\% in velocity estimation.

\subsection{Results of Real-World Experiments}

\begin{figure}[t!]
  \centering
  \subfigure[t][\small{The robot platform.}]{
  \includegraphics[width=0.45\columnwidth]{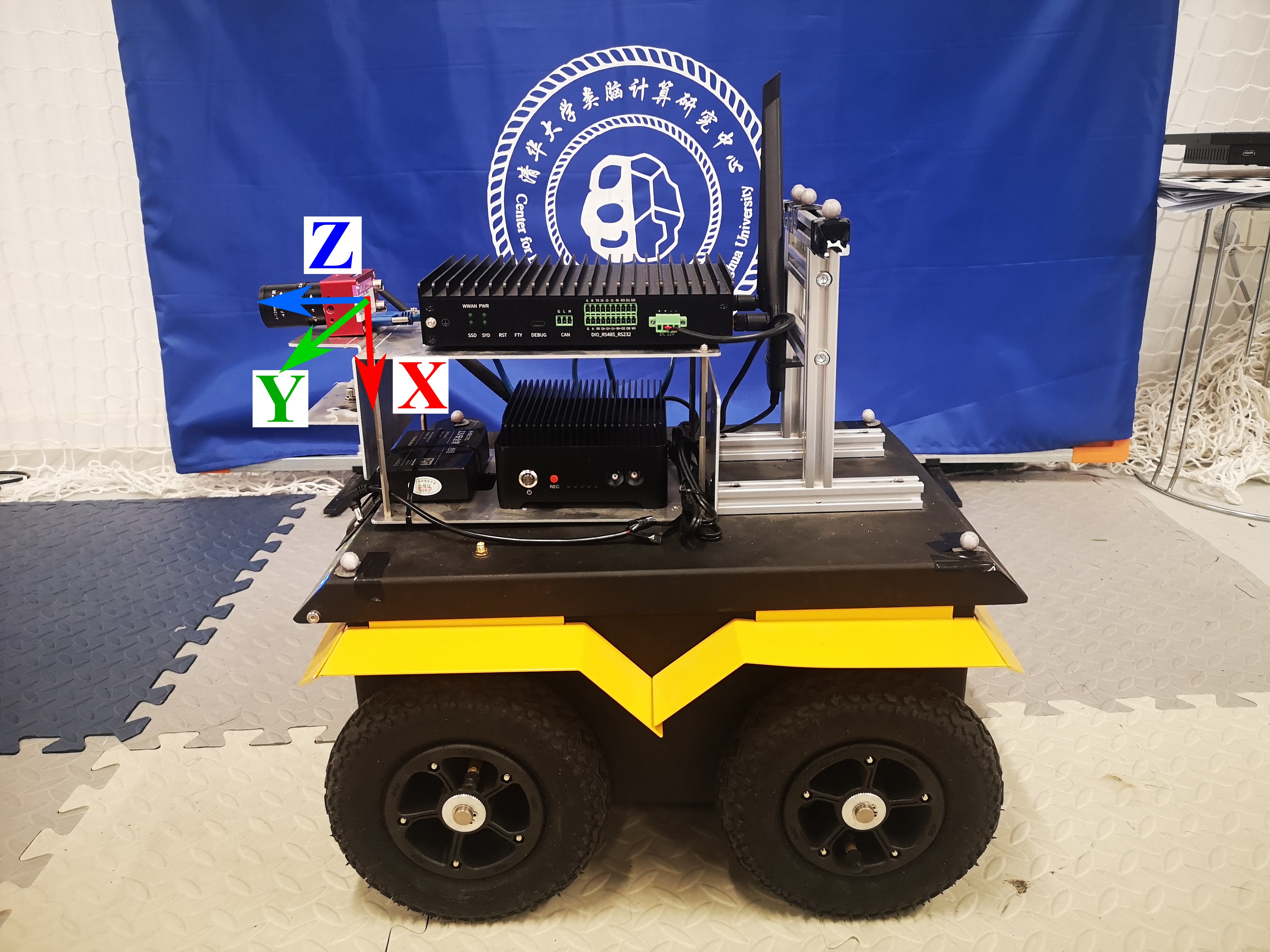}
  \label{subfig: real-world mobile}}\!\!
  \subfigure[t][\small{The scene of real-world experiments.}]{
  \includegraphics[width=0.45\columnwidth]{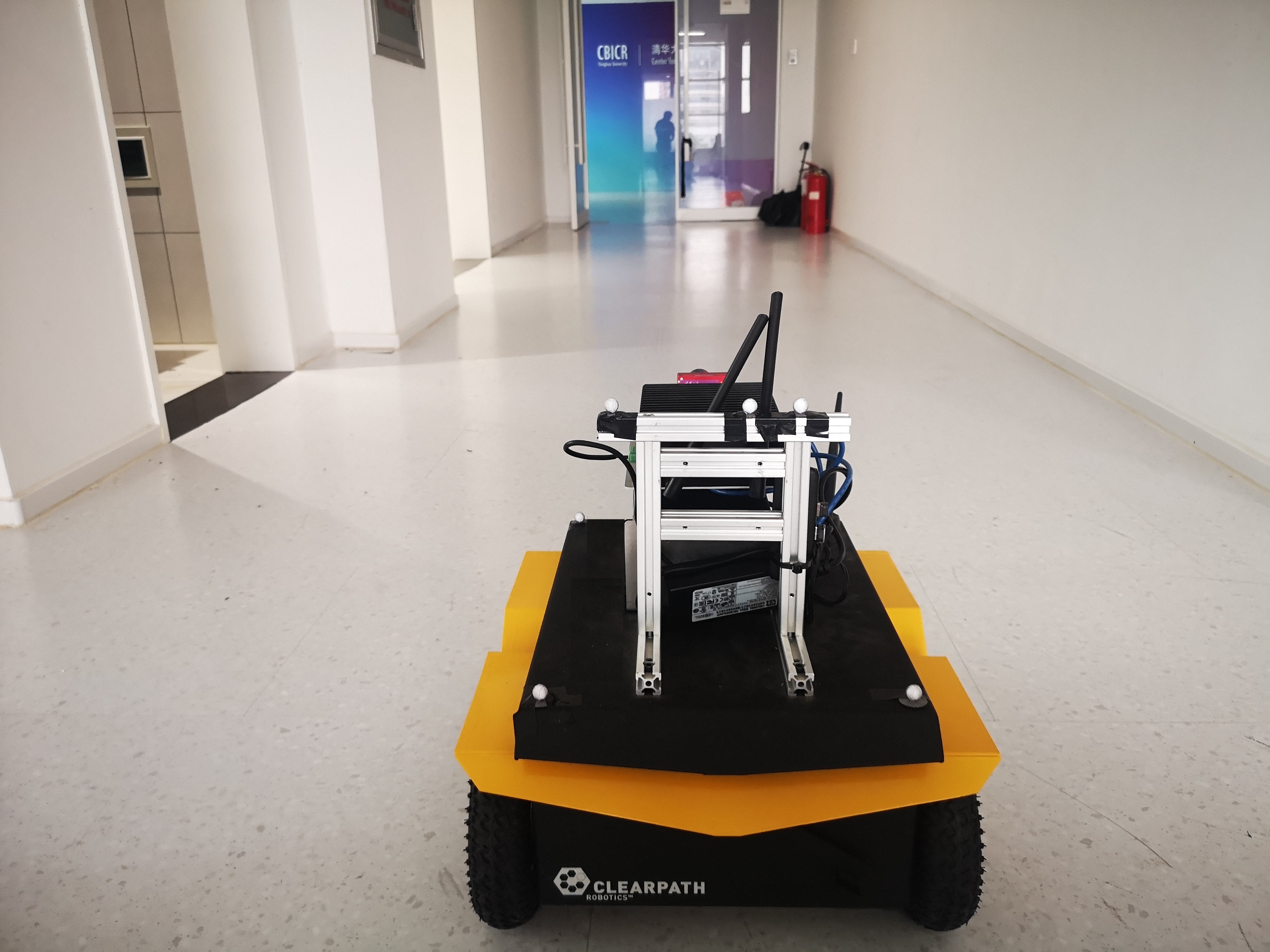}
  \label{subfig: real-world event}}\!\!

  \subfigure[t][\small{Visualization of linear velocity estimation.}]{
  \includegraphics[width=1\columnwidth]{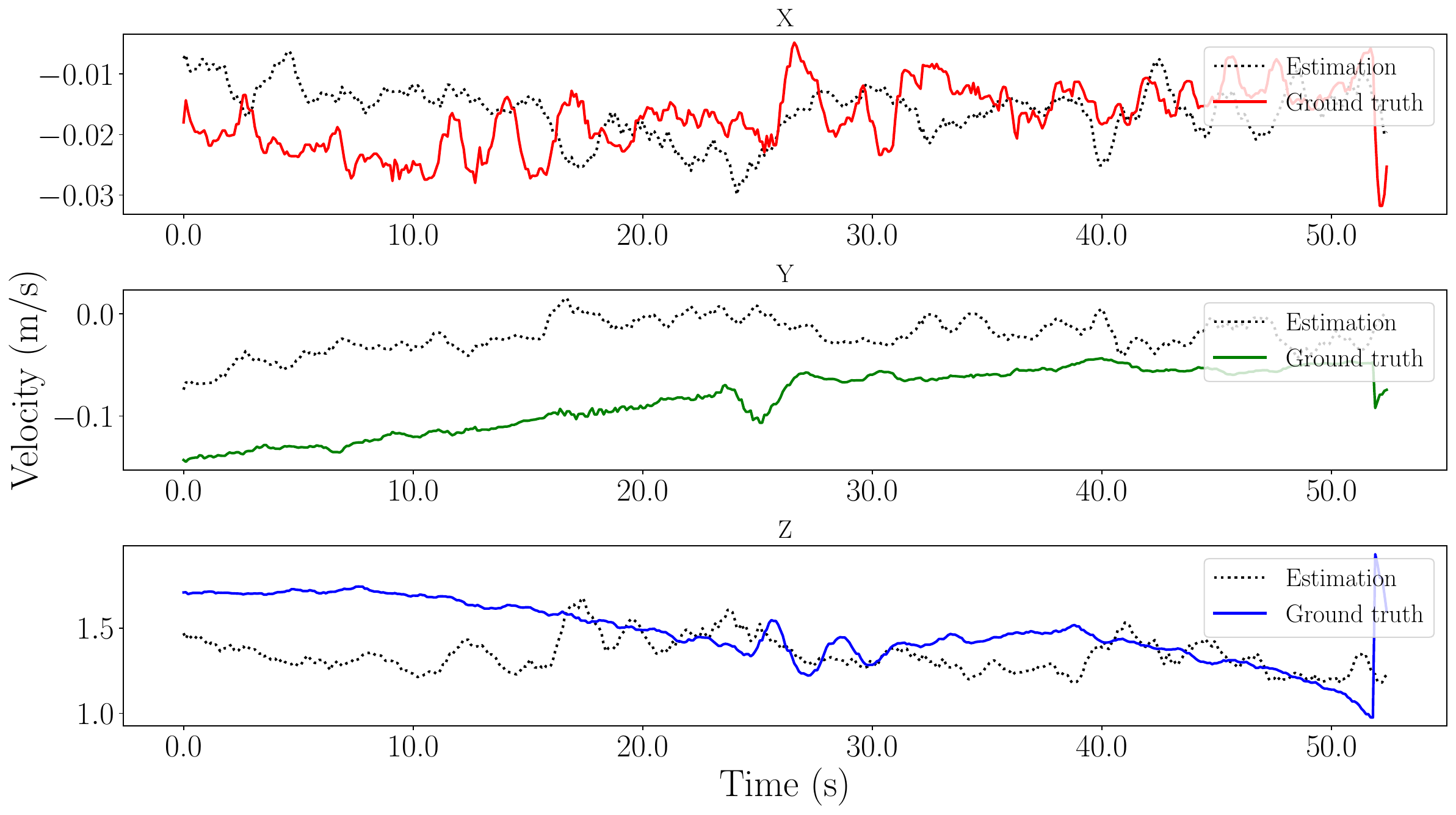}
  \label{subfig: real-world linear}}\!\!

  \caption{Illustration of the robot platform, scene, and results of linear velocity estimation in the real-world experiments.}
  \label{fig: real-world}
  \vspace{-0.5cm}
\end{figure}
Fig.~\ref{subfig: real-world linear} illustrates the estimation of linear velocity.
The coordinate system is built on the camera frame, with the Z-axis aligned directly in front of the camera, the X-axis pointing to the ground, and the Y-axis established according to the right-hand rule.
The experiment's evaluation phase yielded an RMSE of 0.08 and RE of 0.99 for angular velocities, whereas the linear velocities demonstrated an RMSE of 0.21 and RE of 0.23. 
This result exhibits the effectiveness of NeuroVE on real robots.
Furthermore, the vehicle primarily moves in a straight line, rarely requiring steering, which results in angular velocities that are typically much closer to zero. 
Consequently, while the RMSE for angular velocity is smaller when compared to linear velocity, the RE is notably higher for angular velocity.

\section{Conclusion}
\label{sec:conclusion}

This paper introduces NeuroVE, a brain-inspired framework designed to estimate linear and angular velocities by mimicking the vision motion circuit.
It captures the motion data from an event camera and employs mechanisms that mimic the function of LS and AS cells to process the information effectively.
Firstly, we encode temporal information into spikes to simulate time cells in the initial phase.
Subsequently, we introduce ALIF neurons to improve the representation precision of SNNs.
Finally, we introduce the ASLSTM structure incorporated with ALIF neurons to enhance the accuracy of time-series forecasting.
Our method has demonstrated its superior velocity estimation accuracy through synthetic datasets and real-world robot experiments.
Moreover, numerical experiments have substantiated the advantages of NeuroVE in addressing numeric and time-series forecasting issues in SNNs.
The NeuroVE framework provides a novel solution to the essential challenge of ego-velocity estimation, unleashing the potential of neuromorphic computing to address self-motion estimation problems.

\bibliographystyle{IEEEtran} %
% Generated by IEEEtran.bst, version: 1.14 (2015/08/26)

\end{document}